\documentclass[11pt]{article}

\usepackage[top=2cm, bottom=2cm, left=2cm, right=2cm]{geometry}

\usepackage[noadjust]{cite}

\usepackage[cmex10]{amsmath}
\usepackage{amsbsy}

\usepackage{stackengine}
\usepackage{makecell}
\usepackage{boldline}
\setcellgapes{3pt}

\usepackage[latin2]{inputenc}
\usepackage{amsmath}
\usepackage{amsbsy}
\usepackage{bm}
\usepackage{amsthm,dcolumn,graphicx,multicol,fancyhdr}
\usepackage{latexsym}

\usepackage{algorithm}
\usepackage{algpseudocode}
\usepackage{varwidth}

\usepackage{tikz}
\usetikzlibrary{shapes, shadows, arrows}
\usepackage{url}

\usepackage{graphicx}
\usepackage{float}
\usepackage{stfloats}
\usepackage{wrapfig}

\usepackage{epstopdf}

\usepackage{amsthm,dcolumn,graphicx,multicol,fancyhdr}
\usepackage{latexsym}
\usepackage{graphicx}
\usepackage{ifthen}
\usepackage{rotating,amsfonts,color,amssymb,amsmath,amscd,array,enumerate,afterpage}

\usepackage{hyperref}
\usepackage{multirow}

\usepackage{enumitem}
\setlist{parsep=3pt,listparindent=\parindent}

\usepackage[caption=false,font=footnotesize]{subfig}

\usepackage{array}
\newcolumntype{L}[1]{>{\raggedright\let\newline\\\arraybackslash\hspace{0pt}}m{#1}}
\newcolumntype{C}[1]{>{\centering\let\newline\\\arraybackslash\hspace{0pt}}m{#1}}
\newcolumntype{R}[1]{>{\raggedleft\let\newline\\\arraybackslash\hspace{0pt}}m{#1}}

\usepackage{arydshln}

\theoremstyle{definition}

\newcommand{\babc}{\renewcommand{\labelenumi}{(\alph{enumi})}\begin{enumerate}}
\newcommand{\eabc}{\end{enumerate}}
\newcommand{\biii}{\renewcommand{\labelenumi}{(\roman{enumi})}\begin{enumerate}}
\newcommand{\eiii}{\end{enumerate}}

\newcommand{\beqn}{\begin{eqnarray*}}
\newcommand{\beq}{\begin{eqnarray}}
\newcommand{\eeqn}{\end{eqnarray*}}
\newcommand{\eeq}{\end{eqnarray}}

\newcommand{\ckboldon}[1]{#1}

\newcommand{\ckbold}[1]{%
 \ifthenelse{\isundefined{\ckboldon}}{#1}{ \textbf{#1} }
}

\begin{document}
\date{}
\title{Classification of EEG-Based Brain Connectivity Networks in Schizophrenia Using a Multi-Domain Connectome Convolutional Neural Network}
\author{
Chun-Ren Phang\footnote{School of Biomedical Engineering \& Health Sciences, Universiti Teknologi Malaysia, 81310 Skudai, Johor, Malaysia (e-mail: phangcr@gmail.com; cmting@utm.my, mnfuad3@live.utm.my)}, 
Chee-Ming Ting\textsuperscript{*}\textsuperscript{$\dagger$}, 
Fuad Noman\textsuperscript{*},
Hernando Ombao\footnote{Statistics Program, King Abdullah University of Science and Technology, Thuwal 23955, Saudi Arabia (e-mail: hernando.ombao@kaust.edu.sa).}
}

\markboth{IEEE Transactions on Biomedical Engineering}%
{Shell \MakeLowercase{\textit{et al.}}: Bare Demo of IEEEtran.cls for Journals}

\maketitle

\begin{abstract}
\textit{Objective}: We exploit altered patterns in brain functional connectivity as features for automatic discriminative analysis of neuropsychiatric patients. Deep learning methods have been introduced to functional network classification only very recently for fMRI, and the proposed architectures essentially focused on a single type of connectivity measure. \textit{Methods}: We propose a deep convolutional neural network (CNN) framework for classification of electroencephalogram (EEG)-derived brain connectome in schizophrenia (SZ). To capture complementary aspects of disrupted connectivity in SZ, we explore combination of various connectivity features consisting of time and frequency-domain metrics of effective connectivity based on vector autoregressive model and partial directed coherence, and complex network measures of network topology. We design a novel multi-domain connectome CNN (MDC-CNN) based on a parallel ensemble of 1D and 2D CNNs to integrate the features from various domains and dimensions using different fusion strategies. \textit{Results}: Hierarchical latent representations learned by the multiple convolutional layers from EEG connectivity reveal apparent group differences between SZ and healthy controls (HC). Results on a large resting-state EEG dataset show that the proposed CNNs significantly outperform traditional support vector machine classifiers. The MDC-CNN with combined connectivity features further improves performance over single-domain CNNs using individual features, achieving remarkable accuracy of $93.06\%$ with a decision-level fusion. \textit{Conclusion}: The proposed MDC-CNN by integrating information from diverse brain connectivity descriptors is able to accurately discriminate SZ from HC. \textit{Significance}: The new framework is potentially useful for developing diagnostic tools for SZ and other disorders.
\end{abstract}


\vspace{-0.02in}

\vspace{-0.05in}

\section{Introduction}

Schizophrenia (SZ) is considered as a major neuropsychiatric disorder. Neuroimaging studies have provided compelling evidence of both structural functional brain abnormalities in SZ \cite{Shenton2001, Ho2008}. Findings of structural magnetic resonance imaging (MRI) show consistent decrement of brain tissues especially white matter in SZ patients \cite{Peter2005}. Functional MRI (fMRI) studies has associated SZ with functional impairment in sensory and frontal brain areas \cite{Brown2010}. Electrophysiological studies also reported shorten fronto-central distribution of electroencephalogram (EEG) microstates \cite{Kindler2011}, and abnormality of theta and gamma EEG oscillation related to memory impairment of SZ patients \cite{Moran2011}. However, disruption in the function of a single brain region cannot fully explain the range of impairments observed in SZ. There is need to identify altered connectivity between brain regions as a whole network.

Investigating brain connectivity networks based on neuroimaging data has become an important tool to understand the structural and functional organization of the human brain in health and disease. Alterations in the brain connectivity patterns have been associated with various neuropsychiatric disorders and are potentially useful as biomarkers for clinical applications. SZ has been regarded as a dysconnectivity disorder characterized by abnormal structural and functional brain connectivity networks at both microscopic and macroscopic levels \cite{Friston1995,Heuvel2014}. Disconnection of white matter projection tracts has been observed in SZ patients \cite{Skudlarski2010, Zalesky2011, Heuvel2010} using MRI and diffusion tensor imaging (DTI). Analysis of functional connectivity (FC), the statistical dependencies between signals (fMRI or EEG) from spatially distant brain regions, has revealed dysconnectivity in schizophrenic brain networks especially between the frontal regions \cite{Heuvel2014}. EEG studies reported SZ-related aberrant synchronization of neural oscillatory at both the low and high frequencies \cite{Uhlhaas2010, Maran2016}. However, the observed differences in FC patterns between SZ and controls have been inconsistent among fMRI studies, with reports of hyperconnectivity and hypoconnectivity (the increased and decreased strength of connections) between the same brain regions \cite{Skudlarski2010, Liu2008}. There are also contradictory findings from EEG studies reporting both increased \cite{Lehmann2014} and reduced \cite{Tauscher1998, Winterer2001} delta/theta band coherence, and both intact \cite{Tauscher1998, Winterer2001} and reduced \cite{Kam2013} beta-band connectivity in SZ. Complex network analysis based on graph theory \cite{Rubinov2010} has also revealed altered topological organization of brain connectome in SZ patients. Schizophrenic structural brain networks exhibits higher clustering, diminished overall connectivity strength and reduced global efficiency compared to healthy controls \cite{Zalesky2011, Fornito2012}. The network topology of EEG-based connectivity in SZ has not yet been fully investigated.

Machine learning algorithms have been employed for automated classification of altered brain activity in SZ using EEG and fMRI data, primarily based on traditional classifiers such as support vector machine (SVM) \cite{Schnack2014,Huang2018,Kim2016,Sabeti2011,Shim2016,Liu2017}, kernel discriminant analysis (KDA) \cite{Zhu2018} and logistic regression \cite{Sabeti2011}. Moreover, most previous works on EEG-based classification of SZ used time-frequency features from single EEG channels such as band-specific spectral power and univariate autoregression model coefficients \cite{Sabeti2011,Parvinnia2014}. The primary limitation of these single-channel approaches is that they ignore the interactions between channels as a network which has been shown to provide useful information in discriminating SZ from healthy controls at the group-level. Recent advances in deep learning techniques have been shown to be promising for neuroscience applications. \cite{Plis2014}. Deep neural networks (DNNs) have been used to classify brain connectivity of autism spectrum disorder (ASD) with encouraging accuracy \cite{Heinsfeld2018,Kong2019,Guo2017}. Recent studies showed that DNN initialized with deep belief networks outperformed SVM in classifying brain connectome of SZ \cite{Kim2016,Phang2019}.

One popular DNN architecture, the convolutional neural network (CNN), has demonstrated superior capability of representing spatial patterns with remarkable success in classifying two-dimensional (2D) images \cite{krizhevsky2012}. The CNNs, which alternately stack multiple convolutional layers and pooling or sub-sampling layers, can efficiently learn a hierarchy of latent representations that are invariant to small translations of the inputs and allow for parameter sharing via its sparse localized kernels. Applications of CNNs to brain connectome data for classifying spatial maps of functional networks are introduced only recently and in its very early stage. A few related works includes classification of fMRI-derived functional connectivity in mild cognitive impairment \cite{Meszlenyi2017} and in resting-state networks \cite{Zhao2018}, as well as DTI-based structural connectivity for predicting neurodevelopment in infants \cite{Kawahara2017}. Furthermore, the convolutional architecture proposed in these studies focused on one type of connectivity measures from a single domain such as the Pearson correlation for functional connectivity, which do not take into account the connection directionality and topological organization of the brain networks. It remains a challenging task to integrate connectivity metrics from diverse domains (possibly with different dimensions) in the connectome CNNs. To our best knowledge, there are no prior studies using CNN for classification of EEG-based connectivity in SZ. Furthermore, we are not aware of current studies that investigate the classification performance of the combined low-level effective connectivity measures and the global-level complex network measures as input features.

The main contributions of this work are summarized as follows:
\begin{enumerate}[nosep]
	\item We propose a framework based on deep CNNs for classifying altered EEG-derived brain connectivity patterns in SZ.
	\item We exploit, for the first time, effective brain connectivity, a generalization of FC with directionality of information flows between brain regions, as discriminative features in CNN for automatic classification of SZ. We examine various measures of directed connectivity estimated from EEG to capture disrupted brain network organization in SZ. These include the time-domain vector autoregressive (VAR) model coefficients, the frequency-domain partial directed coherence (PDC), the network topology-based complex network (CN) measures, and the combination of these connectivity features.
	\item We design a novel deep CNN architecture called the multi-domain connectome CNN (MDC-CNN) that allows fusion of the time-domain, frequency-domain and the topological measures of brain connectivity networks. The proposed CNN combines an ensemble of 2D-CNN and 1D-CNN in a parallel configuration, taking inputs of VAR (or PDC) and CN connectivity features, respectively. The MDC-CNN is able to learn latent hierarchical representations of the EEG network connectivity pattern with its deep convolutional layers, and then integrate complementary features of the connectivity strength, directionality and network topology to improve classification performance over conventional classifiers.
	\item We examine different fusion strategies in the MDC-CNN in combining CNN classifiers trained on different connectivity feature domains: (1) feature-level fusion by concatenated feature maps of convolutional layers, (2) score-level fusion by concatenated activation outputs of the fully-connected layers, and (3) decision-level fusion by majority voting from independent classifiers, for VAR, PDC and CN features.
\end{enumerate} 

The paper is organized as follows: Section II describes the EEG connectivity features and the proposed MDC-CNN architecture. Section III reports classification performance on a large schizophrenia EEG dataset in comparison with conventional classifiers, for different connectivity feature sets at main EEG frequency bands. Section IV draws the conclusion.

\section{Methods}
\begin{figure*}[!ht]
	\centering
	\begin{minipage}[t]{1\linewidth}
		\centering
		\includegraphics[width=1\linewidth,keepaspectratio]{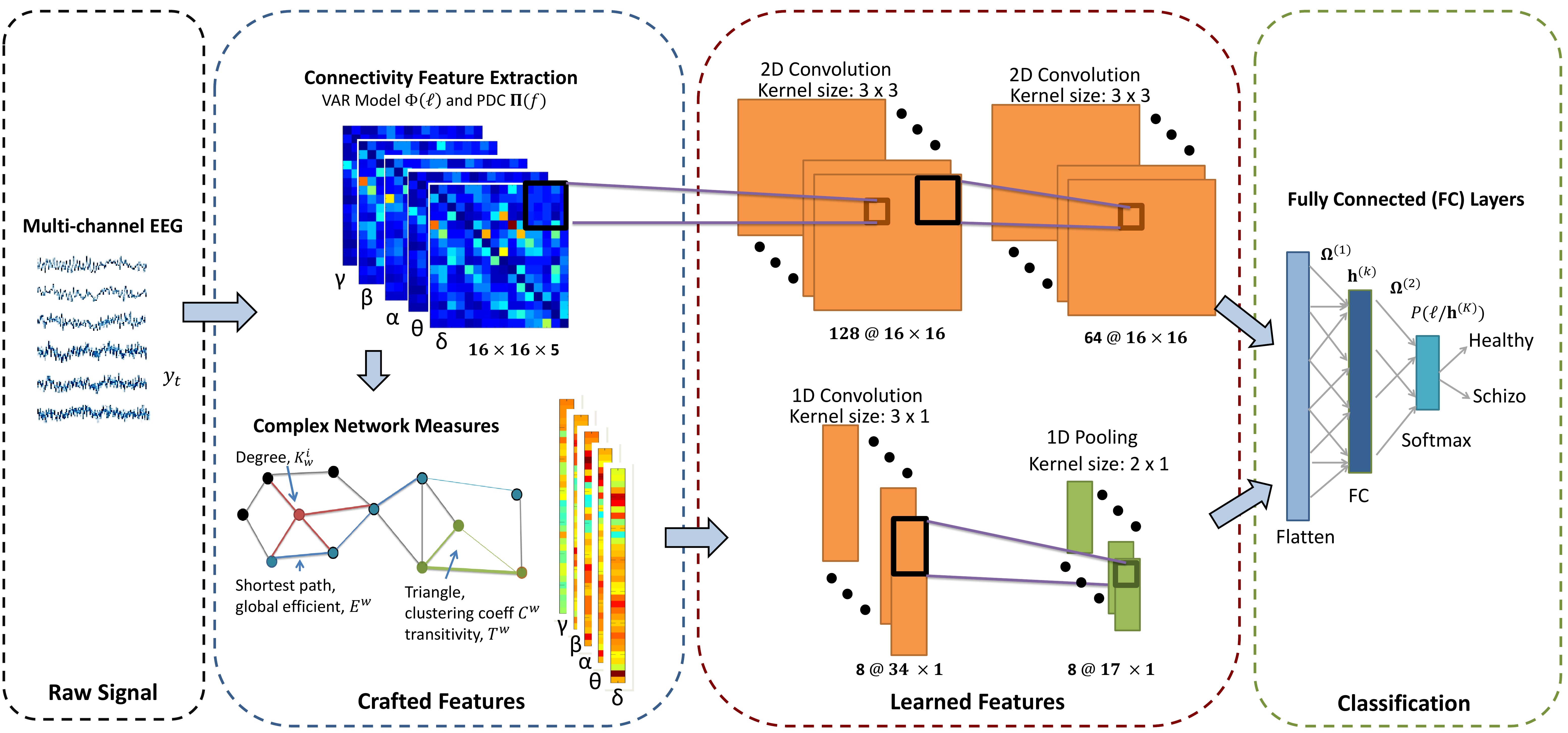}
	\end{minipage}
\vspace{-0.2in}
\caption{Overview of the proposed MDC-CNN-based framework for EEG-derived connectivity pattern classification. Brain connectome features of 2D time-and frequency-domain connectivity matrices and 1D complex network measures are estimated from multi-channel EEG signals. Using these crafted connectivity features in different domain as inputs to an ensemble of deep CNNs in parallel, high-level hierarchical feature maps are then learned by multiple convolutional layers for each domain and combined before passing to classification layers.}
\label{Fig:method}
\end{figure*}

Fig.~\ref{Fig:method} shows an overview of the proposed MDC-CNN framework for classifying SZ and healthy control (HC) using EEG-based effective brain networks, which consists of two stages: Connectivity feature extraction and CNN-based classification. In the first stage, Various crafted measures of directed brain connectivity are estimated from multi-channel EEGs: time-domain VAR coefficients, frequency-domain PDC and topological-based CN measures.
In the second stage, The extracted connectivity features in different domains were then used as inputs to an ensemble of deep CNN classifiers. The two-dimensional connectivity matrices (i.e., VAR coefficient matrices at $L$ lags and PDC matrices at five main frequency bands) are shaped into a 3D-tensor and taken as input to a 2D-CNN model. The one-dimensional vectors of concatenated CN features over five frequency bands are shaped into a 2D-tensor and used as input to a 1D-CNN. The convolution layers of CNN will further learn higher-level features of the spatial structure in the crafted connectivity measures. In fusion of the multiple CNN classifiers, resulting feature maps of different connectivity measures are then flattened and combined, followed by fully-connected layers and a softmax layer for classification into SZ and HC. Alternative strategy is the majority voting of the decisions from independent CNNs trained on different feature domains.

\subsection{Measures of Directed Connectivity}
Let ${\bf y}_t = [y_{1t}, \ldots, y_{Nt}]', t=1, \ldots, T$ be $N$-channel scalp EEG recordings over $T$ time points. A common approach to characterizing effective connectivity between the EEG channels is through a VAR model of order $L$, VAR($L$) on ${\bf y}_t$
\begin{equation} \label{var}
{\bf y}_t = \sum_{\ell=1}^{L} \boldsymbol{\Phi}(\ell) {\bf y}_{t-\ell} + \boldsymbol{\epsilon}_t
\end{equation}
where $\boldsymbol{\epsilon}_t \sim N({\bf 0},\boldsymbol{\Sigma})$ is a white Gaussian noise with mean zero and covariance matrix $\boldsymbol{\Sigma} = E({\boldsymbol{\epsilon}}_t {\boldsymbol{\epsilon}}'_t)$. The directed connectivity network between different EEG channels at time lag $\ell$ is quantified by the $N \times N$ coefficient matrix ${\bf \Phi}({\ell}) = {[\phi_{ij}(\ell)]}_{1 \leq i,j \leq N}$. 

There is said to be the presence of directed influence in a Granger-causality sense from channel $j$ to channel $i$ with strength of $\vert \phi_{ij} \vert$ when $\vert \phi_{ij} \vert > 0$.
Denoting by $\boldsymbol{\beta}=[{\bf \Phi}(1), \ldots, {\bf \Phi}(L)]'$ the VAR coefficients of all lags, model \eqref{var} can be written as a multivariate linear regression
\begin{equation} \label{matvar}
{\bf Y} = {\bf X} \boldsymbol{\beta} + {\bf E}
\end{equation}
where ${\bf Y}=[{\bf y}_{P+1}, \ldots,\ {\bf y}_T]'$, ${\bf E}=[\boldsymbol{\epsilon}_{P+1},\ldots,\boldsymbol{\epsilon}_T]'$ and
\[
{\bf X} =
\left(
  \begin{array}{cccc}
    {\bf y}'_{P}   & {\bf y}'_{P-1} & \ldots & {\bf y}'_{1} \\
		{\bf y}'_{P+1} & {\bf y}'_{P}   & \ldots & {\bf y}'_{2} \\
    \vdots & \vdots & \ldots & \vdots \\
    {\bf y}'_{T-1} & {\bf y}'_{T-2} & \ldots & {\bf y}'_{T-P} \\
  \end{array}
\right)
\]
The estimators of the VAR coefficients can be computed by least-squares (LS) method as $\widehat{\boldsymbol{\beta}} = ({\bf X}' {\bf X})^{-1}{\bf X}'{\bf Y}$ and $\widehat{\boldsymbol{\Sigma}} = (1/(T-L)) ({\bf Y}- {\bf X}\widehat{\boldsymbol{\beta}})' ({\bf Y}- {\bf X}\widehat{\boldsymbol{\beta}})$.

PDC is a frequency-domain measure of effective connectivity which quantifies only the direct dependencies between nodes in a network \cite{Sameshima1999a}. Directed connectivity between EEG channels for oscillatory activity at specific frequency can be characterized by PDC matrix ${\bf \Pi}(f) = {[\pi_{ij}(f)]}_{1 \leq i,j \leq N}$ with
\begin{equation} \label{PDC}
\pi_{ij}(f) = \frac{\left| {\bf \Phi}_{ij}(f) \right|}{\sqrt{\sum_{k=1}^{N} {\left| {\bf \Phi}_{kj}(f) \right|}^2}}
\end{equation}
where ${\bf \Phi}(f) = {\bf I} - \sum_{\ell=1}^{L} {\bf \Phi}(\ell) \text{exp}(-i 2 \pi \ell f/f_s)$ is the Fourier transform of the VAR coefficient matrices with sampling frequency $f_s$. The PDC ${\left| \pi_{ij}(f) \right|}^2 \in [0,1]$ is a normalized measure of the ratio between the outflow of information from channel $y_{jt}$ to $y_{it}$ to the total outflows of all channels from $y_{jt}$ at frequency $f$.

\subsection{Complex Network Analysis}
Complex network (CN) measures derived from graph theory has been widely used to characterize high-order topology of complex brain networks \cite{Rubinov2010}. The CN analysis summarizes large-scale organization of brain networks into neurobiologically meaningful and easily computable measures. It could potentially reveal abnormal functional brain connectivity in psychiatric patients. Four CN measures which reveals important information on disrupted functional integration and segregation in SZ were used to discriminate the brain network structure between SZ and healthy control. Brain network integration can be characterized by the degree and global efficiency; while the clustering coefficient and transitivity detect the network segregation. 

Let ${\bf W} = {[w_{ij}]}$ be $N \times N$ weighted connectivity matrix between EEG channels, which can be quantified by the VAR coefficient matrix $\boldsymbol{\Phi}(\ell)$ at lag $\ell$ or the PDC matrix ${\bf \Pi}(f)$ at specific frequency $f$. The degree $k_{i}^{w}$ of a node $i$ is the total strength of weighted connections to the node, while the global efficiency measures the ease of communication between all nodes in brain network.
\begin{eqnarray}
k_{i}^{w} & = & \sum_{j}{w_{ij}}					\label{degree} \\
E^{w} & = &\frac{1}{N}\sum_{i}{\frac{\sum_{j: j\ne i}{(d_{ij}^{w})^{-1}}}{n-1}}  \label{eff}
\end{eqnarray}
where $d_{ij}^{w}$ is the shortest weighted path length between node $i$ and $j$.

The segregation of a brain network is characterized based on the number of triangles in a network. The clustering coefficient and transitivity quantify the local and global network segregation, given respectively as
\begin{eqnarray}
C^{w} & = & \frac{1}{N}\sum_{i} C_i^{w} = \frac{1}{N}\sum_{i}{\frac{2t_{i}^{w}}{k_{i}(k_{i}-1)}}					\label{cluster} \\
T^{w} & = &\frac{\sum_{i}{2t_{i}^{w}}}{\sum_{i}{k_{i}(k_{i}-1)}}  \label{trans}
\end{eqnarray}
where the clustering coefficient $C^{w}$ of the whole network is the mean of local connectivity clusters $C_i^{w}$ around each node, and the weighted geometric mean of triangles around node $i$ is computed as $t_{i}^{w} = 0.5 \sum_{j,h} {\left(w_{ij}w_{ih}w_{jh}\right)}^{1/3}$.

\subsection{Single-Domain CNN}
In this section, we describe the generic CNN model for classifying individual connectivity feature set from a single-domain, i.e., the VAR coefficient matrices, PDC matrices, or vectors of stacked CN measures. For ease of exposition and without loss of generality, we feed the $N\times N$ PDC matrices ${\bf \Pi}(f)$ over five frequency bands ($\delta, \theta, \alpha, \beta, \gamma $) as input to CNN. CNN is a feed-forward neural network specifically designed to identify patterns in 2D images (or other 1D signals), by incorporating both feature extraction and classification. As in handling 2D-images from different color (RGB) channels in image recognition tasks, CNN can handle the 2D connectivity matrices as images and treat different time-lags or frequency bands as color channels. A typical CNN consists of four layers: convolutional, activation, pooling and fully-connected (or dense) layers.  In the convolutional layer, the input is convolved with a set of $R$ kernels, added by bias terms and mapped through a nonlinear activation function to produce feature maps
\begin{equation} \label{conv-pdc}
\bm{h}_r = g(\sum_{f \in \delta, \theta, \alpha, \beta, \gamma} {\bf \Pi}(f)*\bm{K}_r + \bm{b}_r)
\end{equation} 
where $\bm{K}_r$ is the $r$-th 2D convolution kernel or filter (with dimension $m\times m$) and $\bm{b}_r$ is the bias term for $r = 1, \ldots, R$ with $R$ the number of filters, $*$ denotes the convolution operation in an element-wise form
\begin{equation*}
({\bf \Pi}*\bm{K})_{u,v} = \sum_m \sum_n \pi_{u-m,v-n} K_{u,v}
\end{equation*}
and $g(\cdot)$ is the element-wise activation function.
In subsequent layers, the $q$-th feature map of layer $l$ is connected to $R$ feature maps of the previous layer $l-1$ by
\begin{equation} \label{conv-deep}
\bm{h}_q^{l} = f(\sum_{r=1}^R  \bm{h}_r^{l-1}*\bm{K}_q^{l} + \bm{b}_q^{l})
\end{equation}

The convolution operation offers the advantage of sparse local connectivity where each node of a layer is connected to a small localized area of the inputs. This enables detection of meaningful micro-structure features by using kernels of much smaller size than the input. Convolution layers also allow for parameter sharing where the same weights of the filter are applied across the input map, which prevents overfitting and improves statistical and computational efficiency due to reduced number of network parameters.

Every convolution layer is often followed by a non-learnable layer called pooling, which performs sub-sampling on the feature maps by aggregating small rectangular subsets of values. Max or average pooling is applied to summarize the activation outputs within a rectangular neighborhood with a maximum or an average value. The pooling layers reduce the size of feature maps and enhance invariance of the features to small translations in the input. After several convolution and pooling layers, feature maps are flattened into a feature vector, followed by a fully-connected (FC) layer, plus a final softmax classification layer to output the predictive probabilities of class labels. The parameters of the CNN are trained by minimizing a loss function using gradient descent methods and backpropagation of the error.

\subsection{Proposed MDC-CNN}
\begin{figure*}[!t]
	\centering
	\includegraphics[width=0.9\linewidth,keepaspectratio]{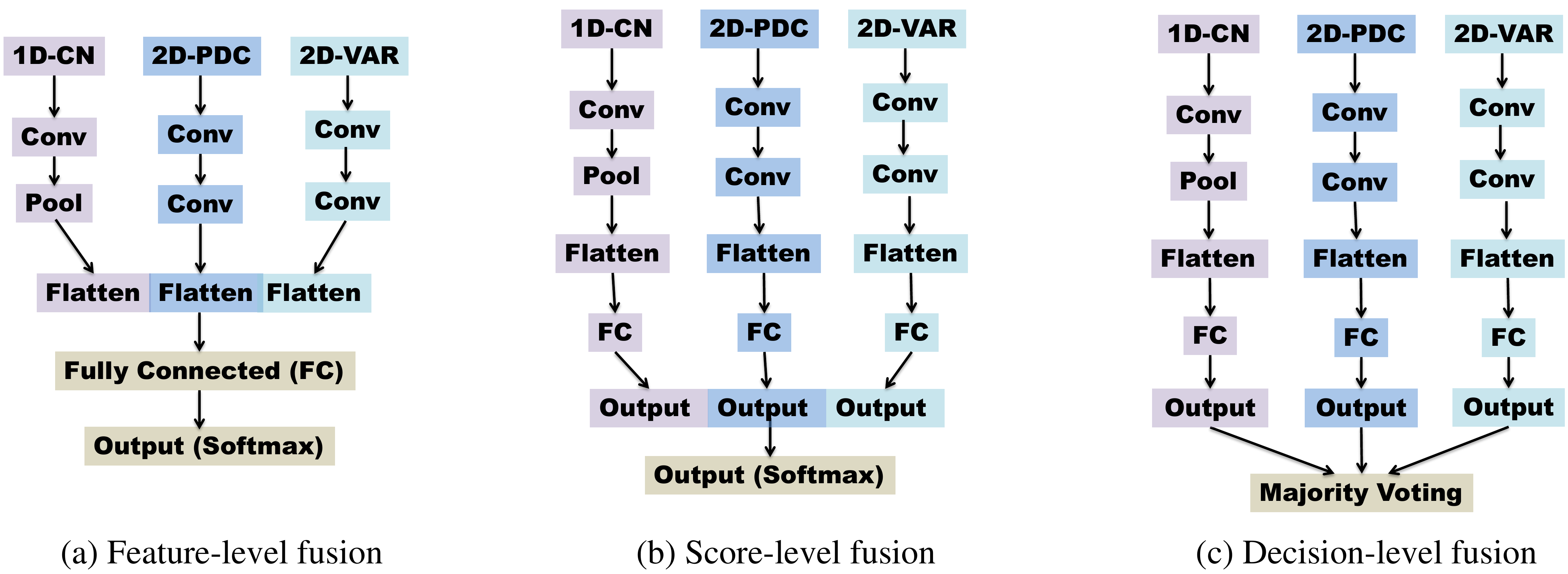}
\vspace{-0.1in}
\caption{Architectures of MDC-CNN model with different fusion strategies in combining multiple domains of connectivity features.}
\label{Fig:MDCCNNs}
\vspace{-0.1in}
\end{figure*}

Brain signals are complex and any analysis should capture the entire brain network using multiple classes of connectivity features. To overcome the limitations of the single-domain CNN, we develop a novel multi-domain connectome CNN model that transforms inputs of low-level VAR and PDC connectivity measures and high-level topological measures of complex brain networks to some latent hierarchical features using 2D-CNN and 1D-CNN, respectively, which are then integrated for classification decision to discriminate SZ and HC. We construct two 2D-CNNs to capture the spatial structure in the two-dimensional VAR and PDC connectivity matrices, respectively and a 1D-CNN for the CN measures of the whole-brain networks. We investigate three different fusion strategies in integrating this ensemble of CNNs trained on multiple domains of connectivity features, as shown in Fig.~\ref{Fig:MDCCNNs}. 
(1) \textbf{Feature-level fusion} (in Fig.~\ref{Fig:MDCCNNs}(a)): Feature maps from a 1D-CNN and two 2D-CNNs based respectively on the CN, VAR and PDC hand-crafted features are flattened and concatenated into a single feature vector, which is then fed into a common classification component consisting of a FC layer and the softmax output layer. In the classification stage, the FC layer approximates a non-linear mapping function which further captures the interactions between the different connectivity features, followed by the softmax activation to yield the class prediction probabilities. 
(2) \textbf{Score-level fusion} (in Fig.~\ref{Fig:MDCCNNs}(b)): Independent CNNs with separate convolutional, FC and softmax layers are learned for individual connectivity feature domains. The predictive probability scores from each of the domain-specific softmax layers are combined as input to a common second-stage softmax layer for predicting the class labels. Additional experiment not reported here showed that this configuration performs better in terms of classifiction accuracy than the conventional score-merging approach by averaging the probability scores over the three CNNs. 
(3) \textbf{Decision-level fusion} (in Fig.~\ref{Fig:MDCCNNs}(c)): The architecture used is similar to that in the score-level fusion. Domain-specific CNNs are trained separately. However, instead of combining the class probability scores, individual output layers of the domain-specific CNNs will predict the class labels independently. This is followed by a majority voting where the final class label is the one that has been predicted most frequently among the three CNN models. Note that the configurations of the convolutional layers are identical for the three connectivity feature sets, and the details will be given in the next section.

\section{Experimental Results}
We evaluate the performance of the proposed MDC-CNN in classifying SZ and HC groups using EEG connectivity patterns measured by VAR, PDC, CN metrics and the fusion of all three feature sets. 
We used a publicly available schizophrenia EEG dataset \cite{SZdata} consisting of 84 subjects (45 SZ patients and 39 HC), provided by the Lomonosov Moscow State University. Resting-state EEG from 16 channels (F7, F3, F4, F8, T3, C3, Cz, C4, T4, T5, P3, Pz, P4, T6, O1 and O2) were recorded with sampling frequency of 128 Hz over duration of 1 minute.

\vspace{-0.02in}
\subsection{Feature Extraction}
Time- and frequency-domain directed connectivity measures (VAR coefficients and PDCs) are estimated from EEG. For each subject, a VAR model was fitted on the EEG signals by LS and PDCs were computed from the estimated VAR coefficients, as in equation \ref{PDC}. The optimal VAR model order selected by the Bayesian information criterion (BIC) averaged over all subjects was $L=5$. Band-limited PDC matrices (self-connections excluded) were computed for five main EEG frequency bands (delta (1-4 Hz), theta (4-7 Hz), alpha (8-13 Hz), beta (14-30 Hz) and gamma (30-64 Hz)). These resulted in 3D-tensor directed connectivity (DC) features ($16 \times 16 \times 5-\text{lags}$ VAR coefficients and $16 \times 16 \times 5-\text{bands}$ PDCs). To quantify the network topology, a set of complex network metrics (16 degrees, 1 global efficiency, 16 clustering coefficients and 1 transitivity) was derived from PDC matrix for each frequency band, giving a total of $34 \times 5$ complex network (CN) features. We compared classification performance of using the time-domain, frequency-domain, topological and the combined connectivity measures as inputs to the proposed MDC-CNN.

\begin{table}[!t]
\caption{Configuration of 2D-CNN (VAR \& PDC connectivity features)}
\vspace{0.2 cm}
\label{Table:CM_archit}
\centering
\resizebox{1\columnwidth}{!}{
\renewcommand{\arraystretch}{0.9}
\begin{tabular}{m{1.5cm}lm{1.5cm}ccccc}
  \hline
Layer & Type & & Number & Size & Stride & Activation & Dropout\\
\hline
1 & Input & & - & $16\times16$ & - & - & 0\\
2 & Convolution & & 128 & $3\times3$ & 1 & ReLU & 0\\
3 & Convolution & & 64 & $3\times3$ & 1 & ReLU & 0\\
4 & Flattening & & - & 16384 & - & - & 0\\
5 & Fully connected & & 64 & - & - & ReLU & 0.5\\
6 & Output & & 2 & - & - & Softmax & 0\\
\hline
\end{tabular}
}
\end{table}

\begin{table}[!t]
\caption{Configuration of 1D-CNN (complex network features)}
\vspace{0.2 cm}
\label{Table:CN_archit}
\centering
\resizebox{1\columnwidth}{!}{
\renewcommand{\arraystretch}{0.9}
\begin{tabular}{m{1.5cm}lm{1.5cm}ccccc}
  \hline
Layer & Type & & Number & Size & Stride & Activation & Dropout\\
\hline
1 & Input & & - & $34\times1$ & - & - & 0\\
2 & Convolution & & 8 & $3\times1$ & 1 & ReLU & 0\\
3 & Average pooling & & - & $2\times1$ & 1 & ReLU & 0\\
4 & Flattening & & - & 136 & - & - & 0\\
5 & Fully connected & & 32 & - & - & ReLU & 0.5\\
6 & Output & & 2 & - & - & Softmax & 0\\
\hline
\end{tabular}
}
\end{table}

\subsection{Model Architecture and Training}
\begin{figure}[!t]
	\centering
	\includegraphics[width=0.6\linewidth,keepaspectratio]{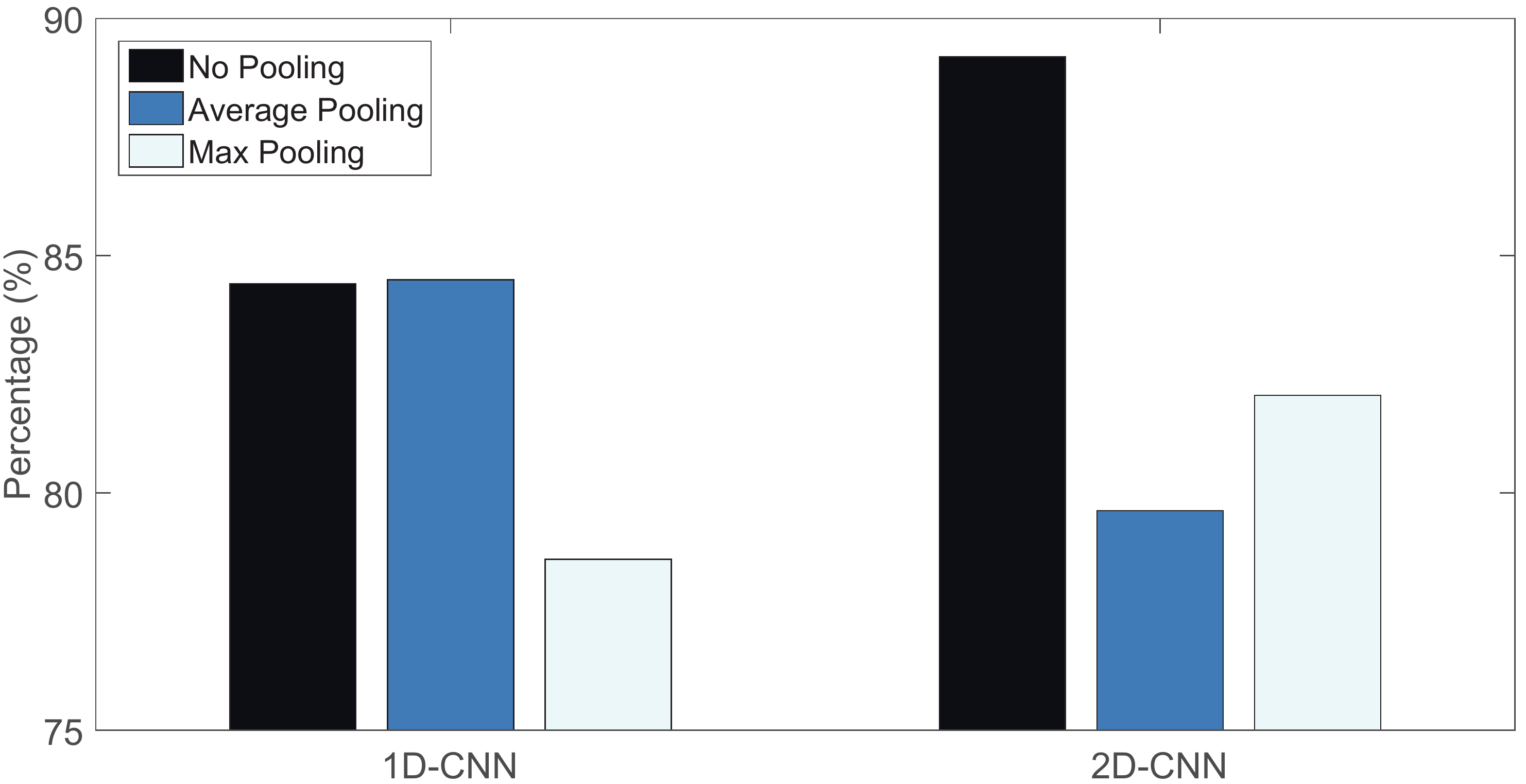}
\vspace{-0.1in}
\caption{Effect of applying pooling layers in the 1D-CNN and 2D-CNN on classification accuracy.}
\label{Fig:pool}
\end{figure}

Here we describe the configurations of the MDC-CNN architecture, choice of hyper-parameters and the tranining procedure. The 2D-CNN was designed to capture spatial structure from the $16\times 16\times 5$ PDC and VAR connectivity matrices across different frequency bands, while the 1D-CNN learns the interactions between CN features. Table~\ref{Table:CM_archit} and Table~\ref{Table:CN_archit} show the architecture of the proposed 2D-CNN and 1D-CNN model, respectively. We omitted the pooling layers in the 2D-CNN in subsequent analysis as it leads to performance drop in classification accuracy compared to without pooling as shown in Fig.~\ref{Fig:pool}. Whereas, adding average pooling in the 1D-CNN improves the performance slightly.

\begin{figure}[!t]
	\centering
	\includegraphics[width=0.5\linewidth,keepaspectratio]{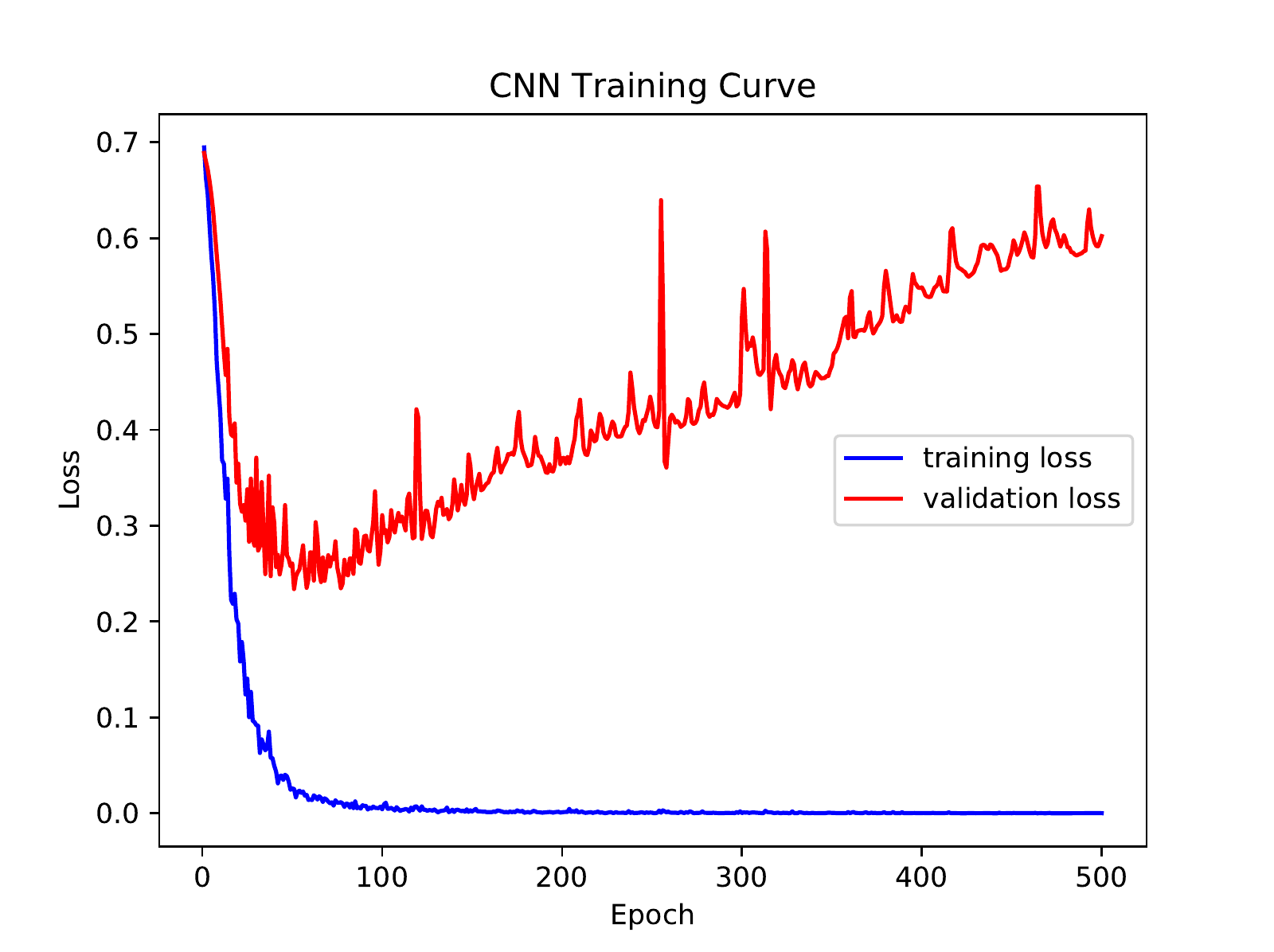}
\vspace{-0.1in}
\caption{The learning curve in cross-entropy loss during the training of the proposed CNN.}
\label{Fig:traincurve}
\end{figure}

\begin{figure}[t!]
	\begin{minipage}[t]{1\linewidth}
		\centering
		\label{Fig:HC1}
		\includegraphics[width=13cm, height=6cm]{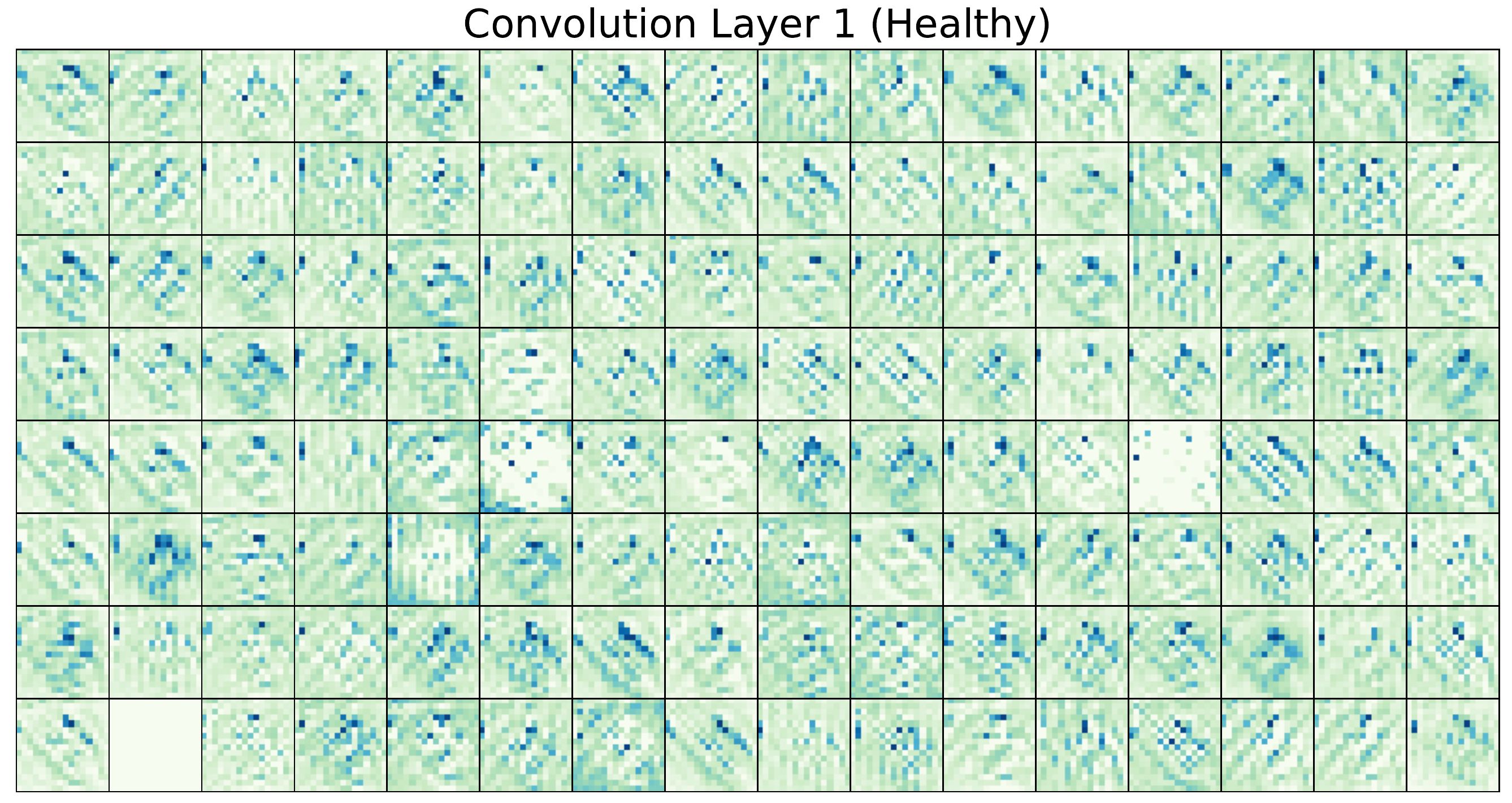}
	\end{minipage}
	\begin{minipage}[t]{1\linewidth}
		\centering
		\label{Fig:SC1}
		\includegraphics[width=13cm, height=6cm]{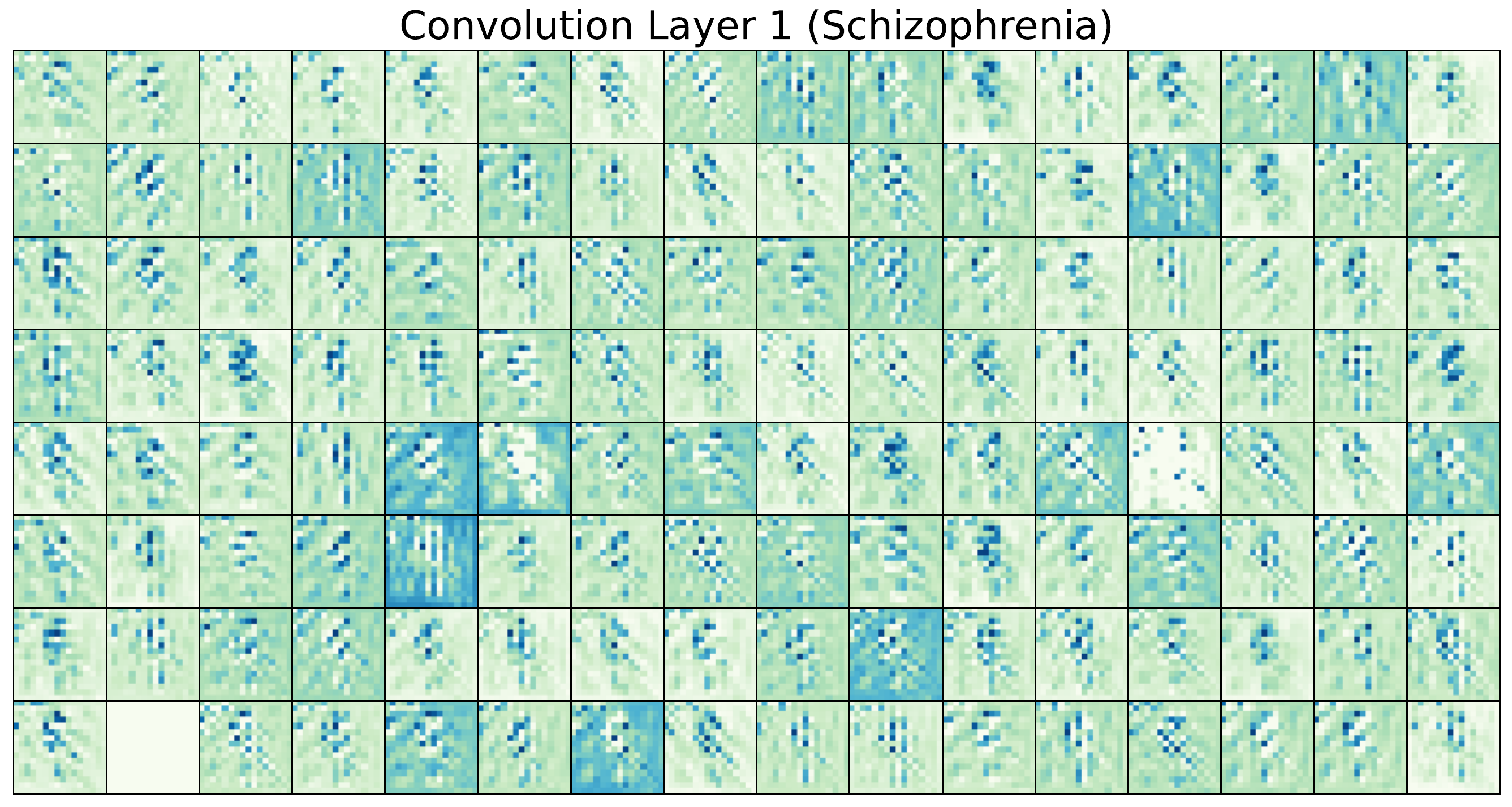}
	\end{minipage}
	\begin{minipage}[t]{0.61\linewidth}
		\centering
		\label{Fig:HC2}
		\includegraphics[width=6cm, height=6cm]{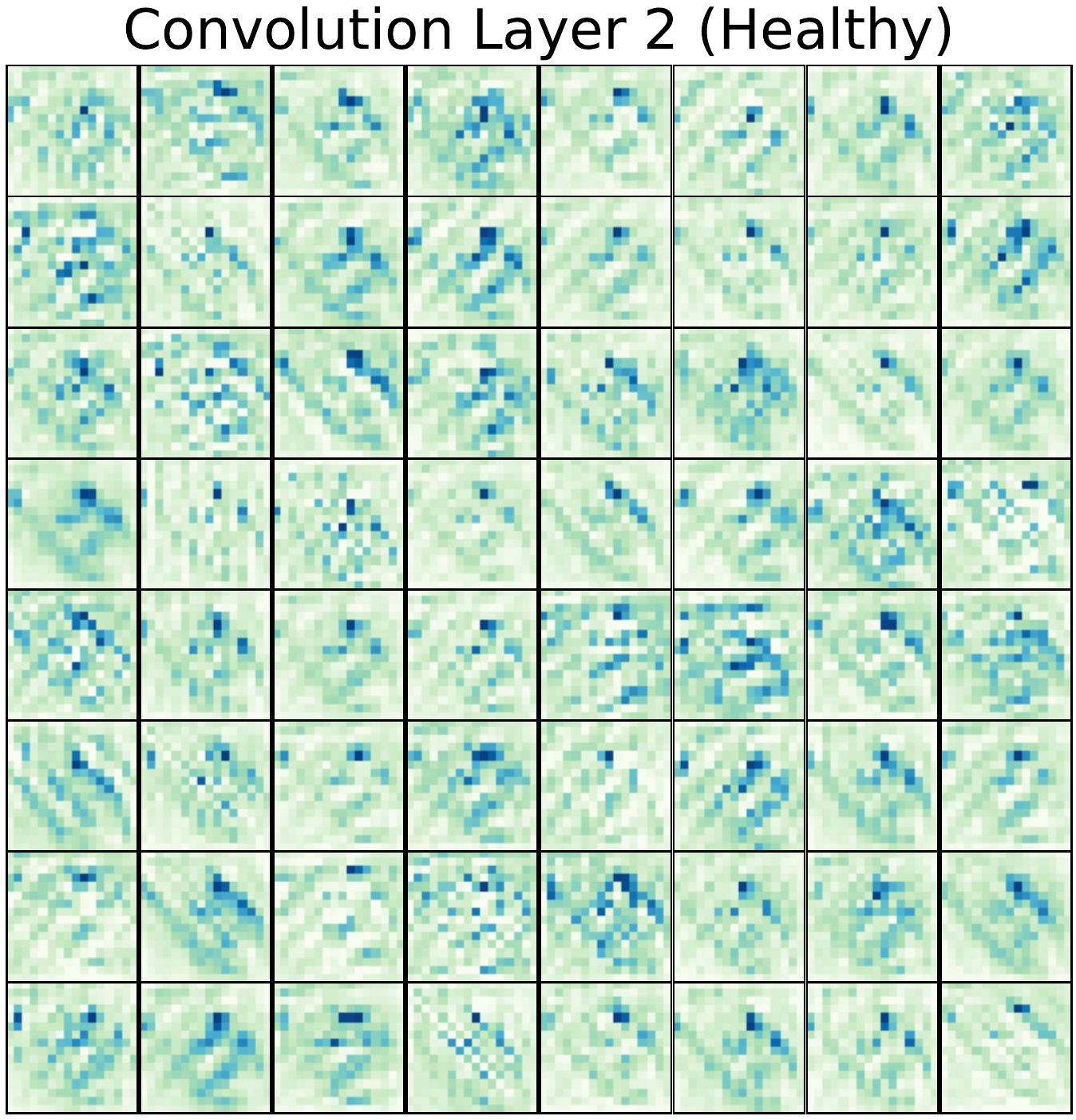}
	\end{minipage}%
	\hspace{-0.65in}
	\begin{minipage}[t]{0.00\linewidth}
		\centering
		\label{Fig:SC2}
		\includegraphics[width=6cm, height=6cm]{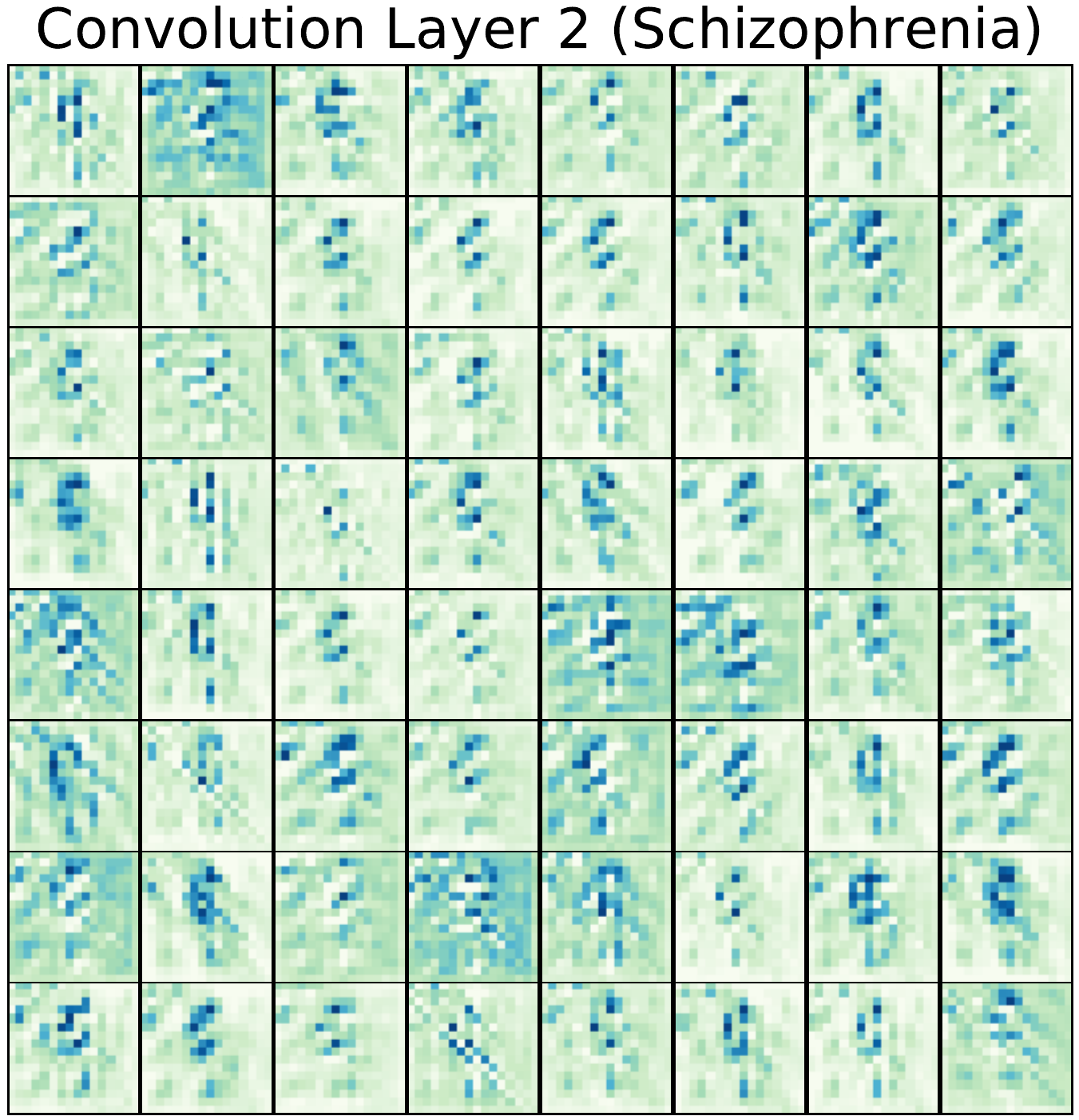}
	\end{minipage}
\caption{Visualization of $16 \times 16$ feature maps of the first and second convolutional layers of the 2D-CNN learned from EEG-derived PDC matrices for SZ and HC subjects. 1st layer: 128 maps. 2nd layer: 64 maps.}
\label{Fig:activation}
\end{figure}

The performance of CNN models were evaluated by five-fold cross validation. The correct classification due to chance on the testing set is 50\%. The extracted connectivity features were used as inputs to train the CNNs, implemented by using Keras \cite{chollet2015}. The CNN was trained using Adam optimizer, with learning rate of 0.0001 and decay of $1e^{-6}$ for 500 epochs. Adam optimizer have been proven to be more efficient in computing stochastic gradient problem \cite{Kingma2014}. 
Dropout with a fixed ratio of 0.5 was applied in the fully-connected layers of the 2D-CNN and 1D-CNN to strengthen the generalization ability of the trained networks \cite{Srivastava2014}. We use rectified linear unit (ReLU) as the activation function $f(x)=\text{max}(0, x)$ for both convolution layers and fully connected layers. To prevent information loss, we applied zero-padding to preserve the shape of features passing through convolution layers. Fig.~\ref{Fig:traincurve} shows the training curve of the CNN models. We selected the optimal model that gives the minimal validation error measured by cross-entropy loss which occur at epoch 80. Cross entropy is defined as the summation of multiplication between class correctness in binary, $b$ and the logarithm probability ($p$) of each class, [$-((b)log(p)+(1-b)log(1-p))$]. The feature maps of the first two convolutional layers of the 2D-CNN learned from the EEG PDC matrices for the SZ and HC subjects are illustrated in Fig.~\ref{Fig:activation}. Apparent difference in the learned feature patterns is found between the two groups with stronger activation outputs of the SZ compared to the HC, particularly evident for some filters at the first layer. This suggests the ability of the proposed model to extract latent features that can discriminate between SZ and controls, and thus improving the classification performance.

\subsection{Classification Results}
Table~\ref{Table:compare} shows the classification performance and standard deviation across subjects of the proposed connectome CNNs using the time-domain VAR, frequency-domain PDC, topological CN features alone in single-domain CNNs, and the fusion of these features by majority voting of decision scores in the MDC-CNN.
As evaluation measures, we used the classification accuracy, sensitivity and specificity. Due to the increased sensitivity to the imbalance between classes, we also computed modified accuracy by averaging the sensitivity and specificity \emph{[Modified accuracy = (sensitivity + specificity) / 2]}. We compared the performance of CNNs with SVM which has shown fairly high accuracy in classifying SZ and HC using features derived from functional connectivity maps in fMRI \cite{Kim2016} and EEG \cite{Aharon2017}.

\begin{table*}[!t]
\caption{Performance comparison of the proposed CNN with SVM using different EEG connectivity features in classifying SZ and HC subjects. Values in parenthesis indicate standard deviations of performance measures}
\vspace{0.3 cm}
\label{Table:compare}
\resizebox{1\columnwidth}{!}{
\renewcommand{\arraystretch}{1.2}
\setlength{\tabcolsep}{1.5pt}
\begin{tabular}{lllcccc}
  \hline
Classifier & Feature & Dimension & Accuracy (\%) & Sensitivity (\%) & Specificity (\%) & Modified Accuracy (\%)\\
\hline
\multirow{4}{*}{SVM} & Time-domain (VAR) & $16\times16\times5(lag)$ & \textbf{88.09}$(\pm3.73)$ & \textbf{87.50} $(\pm7.91)$ & \textbf{88.89} $(\pm9.94)$ & \textbf{88.19}$(\pm3.54)$\\
 & Frequency-domain (PDC) & $16\times16\times5(bands)$ & $80.88 (\pm7.08)$ & $79.64 (\pm14.84)$ & $82.22 (\pm8.89)$ & $80.93 (\pm7.25)$\\
 & Network topology (CN) & $34\times5(bands)$ & $78.60 (\pm6.96)$ & \textbf{87.50} $(\pm13.69)$ & $71.11 (\pm11.33)$ & $79.31 (\pm7.16)$\\
 & Fusion (VAR + PDC + CN) & $((2\times16\times16)+34)\times5$ & $85.74 (\pm4.64)$ & $85.00 (\pm12.25)$ & $86.67 (\pm4.44)$ & $85.83 (\pm5.08)$\\
\hline
\multirow{4}{*}{CNN} & Time-domain (VAR) & $16\times16\times5(lag)$ & $90.59 (\pm7.98)$ & $87.50 (\pm15.81)$ & \textbf{93.33} $(\pm8.89)$ & $90.42 (\pm5.73)$\\
 & Frequency-domain (PDC) & $16\times16\times5(bands)$ & $89.19 (\pm8.83)$ & $84.29 (\pm15.13)$ & \textbf{93.33} $(\pm5.44)$ & $88.81 (\pm6.71)$\\
  & Network topology (CN) & $34\times5(bands)$ & $84.49 (\pm8.87)$ & $81.79 (\pm19.05)$ & $86.67 (\pm10.89)$ & $84.23 (\pm9.28)$\\
 & Fusion (VAR + PDC + CN) & $((2\times16\times16)+34)\times5$ & \textbf{92.87}$(\pm4.38)$ & \textbf{95.00} $(\pm6.12) $ & $91.11 (\pm8.31)$ & \textbf{93.06}$(\pm4.28)$\\
\hline
\end{tabular}}
\end{table*}

\begin{table*}[!t]
\renewcommand{\arraystretch}{1.2}
\caption{Classification performance of CNN using PDC and CN connectivity features for different EEG frequency bands}
\vspace{0.2 cm}
\label{Table:bands}
\centering
\begin{tabular}{llcccc}
  \hline
Feature & Frequency band & Accuracy (\%) & Sensitivity (\%) & Specificity (\%) & Modified Accuracy (\%)\\
\hline
\multirow{5}{*}{PDC} & Delta & $86.84 (\pm7.92)$ & $84.64 (\pm9.29)$ & $88.89 (\pm7.03)$ & $86.77 (\pm8.89)$\\
 & Theta & \textbf{88.01}$(\pm8.48)$ & $81.79 (\pm15.42)$ & \textbf{93.33} $(\pm5.44)$ & \textbf{87.56}$(\pm11.30)$\\
 & Alpha & $86.91 (\pm6.84)$ & $79.64 (\pm9.75)$ & \textbf{93.33} $(\pm5.44)$ & $86.49 (\pm7.12)$\\
 & Beta & $84.56 (\pm5.92)$ & \textbf{87.14} $(\pm13.71)$ & $82.22 (\pm13.33)$ & $84.48 (\pm8.63)$\\
 & Gamma & $84.49 (\pm3.09)$ & $82.50 (\pm10.00)$ & $86.67 (\pm12.96)$ & $84.59 (\pm4.17)$\\
 & All-bands & $89.19 (\pm8.83)$ & $84.29 (\pm15.13)$ & $93.33 (\pm5.44)$ & $88.81 (\pm6.71)$\\
\hline
\multirow{5}{*}{CN} & Delta & $71.25 (\pm10.81)$ & $47.50 (\pm32.02)$ & $91.11 (\pm8.31)$ & $69.31 (\pm8.26)$\\
 & Theta & $77.28 (\pm12.79)$ & $67.14 (\pm26.67)$ & $88.67 (\pm16.33)$ & $77.91 (\pm12.34)$\\
 & Alpha & \textbf{86.99}$(\pm6.79)$ & \textbf{80.00} $(\pm15.00)$ & $93.33 (\pm5.44)$ & \textbf{86.67}$(\pm5.04)$\\
 & Beta & $77.50 (\pm13.51)$ & $56.79 (\pm31.85)$ & \textbf{95.56} $(\pm8.89)$ & $76.18 (\pm14.24)$\\
 & Gamma & $78.46 (\pm8.38)$ & $73.21 (\pm26.63)$ & $82.22 (\pm8.89)$ & $77.72 (\pm10.50)$\\
 & All bands & $84.49 (\pm8.87)$ & $81.79 (\pm19.05)$ & $86.67 (\pm10.89)$ & $84.23 (\pm9.28)$\\
\hline
\end{tabular}
\end{table*}

It is obvious that the CNNs outperformed SVM in classifying SZ and HC EEG connectivity in all feature sets. 
Among the CNN models, the MDC-CNN based on the majority-voting fusion shows superior performance over single-domain CNNs trained on individual connectivity features alone, achieving the best accuracy of 92.87\% and modified accuracy of 93.06\%.
It is interesting to see that time-domain VAR feature performed better than frequency-domain PDC feature, indicating more information are dispersed in time-domain connectivity compared to frequency-domain connectivity. Using the topological CN features alone did not perform as well as other features, with only accuracy of 84.49\% on CNN and 78.60\% on SVM. 
High classification accuracies achieved by the single-domain CNNs suggest that each feature type may capture unique complementary aspects of the brain connectivity networks essential for discriminating HC and SZ. Furthermore, the proposed MDC-CNN integrating features from different domains will significantly improve the classification beyond the achievable performance from each of these feature domains individually.

Table~\ref{Table:bands} shows the classification performance of CNN for different EEG frequency bands. The accuracy of CNN using PDC features ranged from 84.49\% to 88.01\% (with modified accuracy ranged from 84.48\% to 87.56\%), in which the lower frequency bands were more discriminative compared to higher frequency bands. This could be explained by the previous findings that the more pronounced differences in EEG connectivity between the SZ and HC were observed in the low-frequency bands. Since low frequency bands mainly modulate attention, memory and consciousness (for review refer to \cite{Uhlhaas2010}), our results indicate schizophrenia patients are at risk on disruption of above mentioned high-order brain functions.
On the other hand, CN features show lower performance than PDC features in almost all frequency bands, with results in accuracy range of 71.25\% to 86.99\% and modified accuracy of 69.31\% to 86.67\%.
CN features in the alpha band achieve a slightly higher classification accuracy than the PDC, suggesting potentially more serious disruption of the brain network topology at the alpha oscillations. 

\begin{figure}[!t]
	\centering
	\includegraphics[width=0.6\linewidth,keepaspectratio]{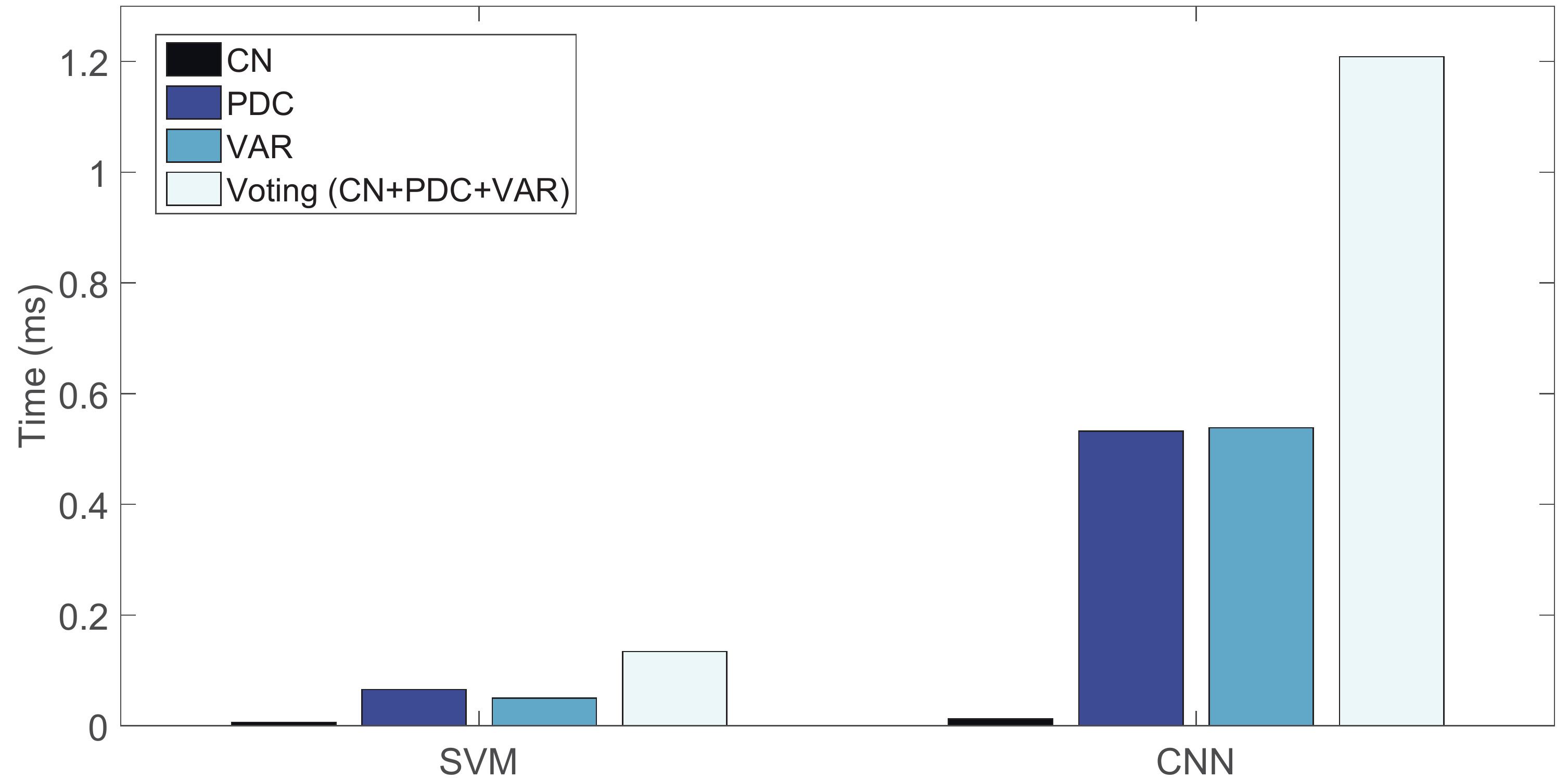}
	\vspace{-0.1in}
\caption{Comparison of the proposed CNN with SVM using different connectivity features in the computational time for classification of a single-subject EEG data.}
\label{Fig:time}
\end{figure}

We also compare the computational time required for the proposed CNN and SVM in classifying different EEG connectivity features from a single subject. The classification was implemented using Intel(R) Core(TM) i7-4790 CPU at 3.60 GHz with 8 GB RAM on 64-bit Windows 7 Professional. As shown in Fig.~\ref{Fig:time}, both classifiers took only fraction of milliseconds (ms) in identifying HC and SZ EEG connectivity. As expected, SVM performed much faster than CNN. However, there is a trade-off between computation time and accuracy.
Although the CNNs are computationally more demanding, our proposed MDC-CNN with majority voting achieved almost 5\% improvement in classification accuracy compared to the best performing SVM (on VAR features), with classification time per subject of only 1.2 ms.

We assess the performance of MDC-CNN under different fusion strategies. The results are shown in Table~\ref{Table:MDCCNNs}. We can see that the decision-level fusion by majority voting outperformed the other two fusion schemes, despite with slightly higher computational cost for classification.

\begin{table}[!t]
\caption{Performance of MDC-CNN for different fusion strategies}
\vspace{0.2 cm}
\label{Table:MDCCNNs}
\centering
\resizebox{0.93\columnwidth}{!}{
\renewcommand{\arraystretch}{1}
\setlength{\tabcolsep}{12pt}
\begin{tabular}{lm{1.5cm}cccc}
  \hline
Performance & & Feature-level & Score-level & Decision-level\\
\hline
Accuracy (\%) & & 91.62 & 91.62 & \textbf{92.87}\\
Sensitivity (\%) & & 90.00 & 89.64 & \textbf{95.00}\\
Specificity (\%) & & \textbf{93.33} & \textbf{93.33} & 91.11\\
Modified Acc. (\%) & & 91.67 & 91.49 & \textbf{93.06}\\
Classification Time (ms) & & 1.19 & 1.04 & 1.2 \\
\hline
\end{tabular}}
\end{table}

\section{Conclusion}
We developed a deep CNN framework for automated classification of SZ patients using input features based on brain connectome signatures derived from EEG. In contrast to the existing approaches, our proposed connectome CNN captures spatial structure of functional connectivity maps, leveraging information on disrupted neural connectivity patterns in SZ relative to the healthy controls for discriminative analysis at the group level.
We introduced the MDC-CNN, a novel CNN architecture designed for multi-domain connectome data that can effectively combine complementary information from multiple brain connectivity descriptors of diverse domains and dimensionality for classification purpose, including time and frequency-domain metrics of effective connectivity and complex network measures of network topology. On resting-state EEG, we demonstrated that the proposed CNN was able to learn a hierarchy of low and high-level abstract representation features from the crafted connectivity features differentiating SZ from HC. The proposed approach gave promising classification results on a large SZ EEG dataset, outperforming traditional SVM by a large margin. The MDC-CNN with combined features also substantially improve the performance over CNNs trained on single-domain features individually, achieving the best modified accuracy of $93.06\%$ using decision-level fusion. Our framework is generally applicable to other neuropsychiatric disorders besides SZ associated with aberrant connectivity patterns and is potentially useful for development of robust computer-aided diagnostic tools in clinical settings. Future work will consider extensions to incorporate time-varying or dynamic effective connectivity metrics \cite{Samdin2017,Ting2018} as additional features to enhance classification performance. A dynamic classification framework \cite{Park2018} could be implemented from time-varying connectivity features. Sequential deep learning models such as the recurrent neural networks (RNNs) or long short-term memory (LSTM) RNNs could be used to capture the temporal dependencies in the dynamic brain networks \cite{Wang2018,WangY2018}.

\newpage

\vspace{-0.1in}
\bibliographystyle{IEEEbib_2names}
\bibliography{ref}

\begin{thebibliography}{10}

\bibitem{Shenton2001}
M.~E. Shenton, et~al.,
\newblock ``A review of {MRI} findings in schizophrenia,''
\newblock {\em Schizophr. Res.}, vol. 49, no. 1-2, pp. 1--52, 2001.

\bibitem{Ho2008}
H.~M., et~al.,
\newblock ``Time-frequency discriminant analysis of meg signals,''
\newblock {\em NeuraImage}, vol. 40, no. 1, pp. 174--186, 2008.

\bibitem{Peter2005}
F.~B. Peter,
\newblock ``Neuroimaging of schizophrenia: Structural abnormalities and
  pathophysiological implications,''
\newblock {\em Neuropsychiatr. Dis. Treat.}, vol. 1, no. 3, pp. 193--204, 2005.

\bibitem{Brown2010}
G.~G. Brown and W.~K. Thompson,
\newblock {\em Functional Brain Imaging in Schizophrenia: Selected Results and
  Methods}, pp. 181--214,
\newblock Springer Berlin Heidelberg, Berlin, Heidelberg, 2010.

\bibitem{Kindler2011}
J.~Kindler, et~al.,
\newblock ``Resting-state {EEG} in schizophrenia: Auditory verbal
  hallucinations are related to shortening of specific microstates,''
\newblock {\em Clin. Neurophysiol.}, vol. 122, no. 6, pp. 1179--1182, 2011.

\bibitem{Moran2011}
L.~V. Moran and L.~E. Hong,
\newblock ``High vs low frequency neural oscillations in schizophrenia,''
\newblock {\em Schizophr. Bull.}, vol. 37, no. 4, pp. 659--663, 2011.

\bibitem{Friston1995}
K.~J. Friston and C.~D. Frith,
\newblock ``Schizophrenia: A disconnection syndrome,''
\newblock {\em Clin. Neurosci.}, vol. 3, no. 2, pp. 89--97, 1995.

\bibitem{Heuvel2014}
M.~P. van~den Heuvel and A.~Fornito,
\newblock ``Brain networks in schizophrenia,''
\newblock {\em Neuropsychol. Rev.}, vol. 24, no. 1, pp. 32--48, 2014.

\bibitem{Skudlarski2010}
P.~Skudlarski, et~al.,
\newblock ``Brain connectivity is not only lower but different in
  schizophrenia: A combined anatomical and functional approach,''
\newblock {\em Biol. Psychiatry}, vol. 68, no. 1, pp. 61--69, 2010.

\bibitem{Zalesky2011}
A.~Zalesky, et~al.,
\newblock ``Disrupted axonal fiber connectivity in schizophrenia,''
\newblock {\em Biol. Psychiatry}, vol. 69, no. 1, pp. 80--89, 2011.

\bibitem{Heuvel2010}
M.~P. van~den Heuvel, et~al.,
\newblock ``Aberrant frontal and temporal complex network structure in
  schizophrenia: A graph theoretical analysis,''
\newblock {\em J. Neurosci.}, vol. 30, no. 47, pp. 15915--15926, 2010.

\bibitem{Uhlhaas2010}
P.~J. Uhlhaas and W.~Singer,
\newblock ``Abnormal neural oscillations and synchrony in schizophrenia,''
\newblock {\em Nat. Rev. Neurosci.}, vol. 11, no. 2, pp. 100, 2010.

\bibitem{Maran2016}
M.~Maran, et~al.,
\newblock ``Electrophysiological insights into connectivity anomalies in
  schizophrenia: {A} systematic review,''
\newblock {\em Neuropsychiatr. Electrophysiol.}, vol. 2, no. 1, pp. 6, 2016.

\bibitem{Liu2008}
Y.~Liu, et~al.,
\newblock ``Disrupted small-world networks in schizophrenia,''
\newblock {\em Brain}, vol. 131, no. 4, pp. 945--961, 2008.

\bibitem{Lehmann2014}
D.~Lehmann, et~al.,
\newblock ``Functionally aberrant electrophysiological cortical connectivities
  in first episode medication-naive schizophrenics from three psychiatry
  centers,''
\newblock {\em Front. Hum. Neurosci.}, vol. 8, pp. 635, 2014.

\bibitem{Tauscher1998}
J.~Tauscher, et~al.,
\newblock ``Low frontal electroencephalographic coherence in neuroleptic-free
  schizophrenic patients,''
\newblock {\em Biol. Psychiatry}, vol. 44, no. 6, pp. 438--447, 1998.

\bibitem{Winterer2001}
G.~Winterer, et~al.,
\newblock ``An association between reduced interhemispheric {EEG} coherence in
  the temporal lobe and genetic risk for schizophrenia,''
\newblock {\em Schizophr. Res.}, vol. 49, no. 1-2, pp. 129--143, 2001.

\bibitem{Kam2013}
J.~W. Kam, et~al.,
\newblock ``{Resting state {EEG} power and coherence abnormalities in bipolar
  and schizophrenia},''
\newblock {\em J. Psychiatr. Res.}, vol. 47, no. 12, pp. 1893--1901, 2013.

\bibitem{Rubinov2010}
M.~Rubinov and O.~Sporns,
\newblock ``{Complex network measures of brain connectivity: Uses and
  Interpretations},''
\newblock {\em NeuroImage}, vol. 52, no. 3, pp. 1059--1069, 2010.

\bibitem{Fornito2012}
A.~Fornito, et~al.,
\newblock ``Schizophrenia, neuroimaging and connectomics,''
\newblock {\em NeuroImage}, vol. 62, no. 4, pp. 2296--2314, 2012.

\bibitem{Schnack2014}
H.~G. Schnack, et~al.,
\newblock ``Can structural {MRI} aid in clinical classification? a machine
  learning study in two independent samples of patients with schizophrenia,
  bipolar disorder and healthy subjects,''
\newblock {\em NeuroImage}, vol. 84, pp. 299--306, 2014.

\bibitem{Huang2018}
J.~Huang, et~al.,
\newblock ``Identifying resting-state multi-frequency biomarkers via
  tree-guided group sparse learning for schizophrenia classification,''
\newblock {\em IEEE J. Biomed. Heal. Informatics}, pp. 1--1, 2018.

\bibitem{Kim2016}
J.~Kim, et~al.,
\newblock ``Deep neural network with weight sparsity control and pre-training
  extracts hierarchical features and enhances classification performance:
  Evidence from whole-brain resting-state functional connectivity patterns of
  schizophrenia,''
\newblock {\em NeuroImage}, vol. 124, pp. 127--146, 2016.

\bibitem{Sabeti2011}
M.~Sabeti, et~al.,
\newblock ``A new approach for {EEG} signal classification of schizophrenic and
  control participants,''
\newblock {\em Expert Syst. Appl.}, vol. 38, no. 3, pp. 2063--2071, 2011.

\bibitem{Shim2016}
M.~Shim, et~al.,
\newblock ``Machine-learning-based diagnosis of schizophrenia using combined
  sensor-level and source-level {EEG} features,''
\newblock {\em Schizophr. Res.}, vol. 176, no. 2, pp. 314--319, 2016.

\bibitem{Liu2017}
H.~Liu, et~al.,
\newblock ``A data driven approach for resting-state {EEG} signal
  classification of schizophrenia with control participants using random matrix
  theory,''
\newblock {\em ArXiv e-prints}, 2017.

\bibitem{Zhu2018}
Q.~Zhu, et~al.,
\newblock ``Non-negative discriminative brain functional connectivity for
  identifying schizophrenia on resting-state f{MRI},''
\newblock {\em Biomed. Eng. Online.}, vol. 17, no. 1, pp. 32, 2018.

\bibitem{Parvinnia2014}
E.~Parvinnia, et~al.,
\newblock ``Classification of {EEG} signals using adaptive weighted distance
  nearest neighbor algorithm,''
\newblock {\em J. King Saud Univ. - Comput. Inf. Sci.}, vol. 26, no. 1, pp.
  1--6, 2014.

\bibitem{Plis2014}
S.~M. Plis, et~al.,
\newblock ``Deep learning for neuroimaging: {A} validation study,''
\newblock {\em Front. Neurosci.}, vol. 8, pp. 229, 2014.

\bibitem{Heinsfeld2018}
A.~S. Heinsfeld, et~al.,
\newblock ``Identification of autism spectrum disorder using deep learning and
  the abide dataset,''
\newblock {\em NeuroImage Clin.}, vol. 17, pp. 16--23, 2018.

\bibitem{Kong2019}
Y.~Kong, et~al.,
\newblock ``Classification of autism spectrum disorder by combining brain
  connectivity and deep neural network classifier,''
\newblock {\em Neurocomputing}, vol. 324, pp. 63--68, 2019.

\bibitem{Guo2017}
X.~Guo, et~al.,
\newblock ``Diagnosing autism spectrum disorder from brain resting-state
  functional connectivity patterns using a deep neural network with a novel
  feature selection method,''
\newblock {\em Front. Neurosci.}, vol. 11, no. 460, 2017.

\bibitem{Phang2019}
C.-R. Phang, et~al.,
\newblock ``Classification of {EEG}-based effective brain connectivity in
  schizophrenia using deep neural networks,''
\newblock in {\em Proc. 2019 IEEE Int. Conf. Neural Eng.}, 2019.

\bibitem{krizhevsky2012}
A.~Krizhevsky, et~al.,
\newblock ``Imagenet classification with deep convolutional neural networks,''
\newblock in {\em Adv. Neural Inf. Process. Syst.}, 2012, pp. 1097--1105.

\bibitem{Meszlenyi2017}
R.~J. Meszl{\'e}nyi, et~al.,
\newblock ``Resting state f{MRI} functional connectivity-based classification
  using a convolutional neural network architecture,''
\newblock {\em Front. Neuroinformatics}, vol. 11, pp. 61, 2017.

\bibitem{Zhao2018}
Y.~Zhao, et~al.,
\newblock ``Automatic recognition of f{MRI}-derived functional networks using
  3-{D} convolutional neural networks,''
\newblock {\em IEEE Trans. Biomed. Eng.}, vol. 65, no. 9, pp. 1975--1984, 2018.

\bibitem{Kawahara2017}
J.~Kawahara, et~al.,
\newblock ``Brain{N}et{CNN}: Convolutional neural networks for brain networks;
  towards predicting neurodevelopment,''
\newblock {\em NeuroImage}, vol. 146, pp. 1038--1049, 2017.

\bibitem{Sameshima1999a}
K.~Sameshima and L.~A. Baccala,
\newblock ``{Using partial directed coherence to describe neuronal ensemble
  interactions},''
\newblock {\em J. Neurosci. Methods}, vol. 94, no. 1, pp. 93--103, 1999.

\bibitem{SZdata}
N.~N. Gorbachevskaya and S.~Borisov,
\newblock ``{EEG} data of healthy adolescents and adolescents with symptoms of
  schizophrenia,'' \url{http://brain.bio.msu.ru/eeg_schizophrenia.htm}.

\bibitem{chollet2015}
C.~Fran\c{c}ois et~al.,
\newblock ``Keras,'' \url{https://keras.io}, 2015.

\bibitem{Kingma2014}
D.~P. Kingma and J.~Ba,
\newblock ``Adam: A method for stochastic optimization,''
\newblock {\em CoRR}, vol. abs/1412.6980, 2014.

\bibitem{Srivastava2014}
N.~Srivastava, et~al.,
\newblock ``{Dropout: A Simple Way to Prevent Neural Networks from
  Overfitting},''
\newblock {\em J. Mach. Learn. Res.}, vol. 15, pp. 1929--1958, 2014.

\bibitem{Aharon2017}
Z.~Dvey-Aharon, et~al.,
\newblock ``Connectivity maps based analysis of {EEG} for the advanced
  diagnosis of schizophrenia attributes,''
\newblock {\em PLOS ONE}, vol. 12, no. 10, pp. 1--15, 2017.

\bibitem{Samdin2017}
S.~B. Samdin, et~al.,
\newblock ``A unified estimation framework for state-related changes in
  effective brain connectivity,''
\newblock {\em IEEE Trans. Biomed. Eng.}, vol. 64, no. 4, pp. 844--858, 2017.

\bibitem{Ting2018}
C.-M. Ting, et~al.,
\newblock ``Estimating dynamic connectivity states in f{MRI} using
  regime-switching factor models,''
\newblock {\em IEEE Trans. Med. Imaging}, vol. 37, no. 4, pp. 1011--1023, 2018.

\bibitem{Park2018}
T.~Park, et~al.,
\newblock ``Dynamic classification using multivariate locally stationary
  wavelet processes,''
\newblock {\em Signal Processing}, vol. 152, pp. 118 -- 129, 2018.

\bibitem{Wang2018}
H.~Wang, et~al.,
\newblock ``Recognizing brain states using deep sparse recurrent neural
  network,''
\newblock {\em IEEE Trans. Med. Imag.}, 2018.

\bibitem{WangY2018}
Y.~Wang, et~al.,
\newblock ``Estimating brain connectivity with varying-length time lags using a
  recurrent neural network,''
\newblock {\em IEEE Trans. Biomed. Eng.}, vol. 65, no. 9, pp. 1953--1963, 2018.

\end{thebibliography}

\end{document}